\documentclass[journal]{IEEEtran}
\usepackage{graphicx}
\usepackage{amssymb}
\usepackage{subfigure}
\usepackage{setspace}
\usepackage{amsmath}
\usepackage{caption2}

\begin{document}
 \title{Face Parsing via a Fully-Convolutional Continuous CRF Neural Network}
 \author{Lei Zhou, Zhi Liu, Senior Member, IEEE, Xiangjian He, Senior Member, IEEE}
\maketitle

  \begin{abstract}
    In this work, we address the face parsing task with a Fully-Convolutional continuous CRF Neural Network (FC-CNN) architecture.
    In contrast to previous face parsing methods that apply region-based subnetwork hundreds of times, our FC-CNN is fully convolutional with high segmentation accuracy. To achieve this goal, FC-CNN integrates three subnetworks, a unary network, a pairwise network and a continuous Conditional Random Field (C-CRF) network into a unified framework. The high-level semantic information and low-level details across different convolutional layers are captured by the convolutional and deconvolutional structures in the unary network. The semantic edge context is learnt by the pairwise network branch to construct pixel-wise affinity. Based on a differentiable superpixel pooling layer and a differentiable C-CRF layer, the unary network and pairwise network are combined via a novel continuous CRF network to achieve spatial consistency in both training and test procedure of a deep neural network. Comprehensive evaluations on LFW-PL and HELEN datasets demonstrate that FC-CNN achieves better performance over the other state-of-arts for accurate face labeling on challenging images.
  \end{abstract}

 %
\IEEEpeerreviewmaketitle
 \section{Introduction}
  \indent  The task of face parsing is to assign a categorical label to every pixel in a face image. It enables many high level applications, e.g, hair editing and face beautification. While there has been previous work related to face parsing based on landmark representations \cite{LANDMARK1}, conditional random field \cite{CRFFACE1,CRFFACE2} and exemplar \cite{EXEMPLAR}, none of the previous methods has achieved an excellent dense prediction over raw image pixels in a fully end-to-end way.\\
  \indent Over the last few years, the success of deep convolutional neural network (CNN) models \cite{alexnet,verydeepcnn,he2016deep}
  has resulted in a dramatic progress in the task of pixle-wise semantic segmentation using rich hierarchical features \cite{FCN,SEGNET,CRFFCN,DEEPLABDT,DEEPV1,DEEPRFT}. Most of the current semantic segmentation methods are derived from a fully convolutional network (FCN), which was first introduced in \cite{FCN}. In FCN, the last few fully connected layers are replaced by a convolutional layer to make efficient an end-to-end learning and inference.\\
   \indent A dominant research direction for improving semantic segmentation with deep learning
   is the combination of the powerful classification capabilities of FCNs with a structured prediction,
   which aims at improving classification by capturing the interactions between predicted labels. Probabilistic graph
   models have been popular for a long time for structured prediction of labels, with constraints enforcing label consistency.
   The conditional random field (CRF) is a common framework which utilizes both the local and global dependencies
    within an image to refine the prediction map. Various models \cite{AHC,COOS,WWH} based on higher order clique potentials have
    been developed to improve the segmentation performance. Most current state-of-the-art methods \cite{DEEPV1,LIN,OCP,CRFFCN}
    have incorporated graphical models into a deep learning framework. One of the first work of combining deep learning
    framework with a structured prediction was proposed in \cite{DEEPV1} and it applied the densely connected conditional random field
     (DenseCRF) \cite{fullyc} to the post-processes of FCN output to generate a better segmentation with refined image boundaries. Zheng et al. \cite{CRFFCN} combined DenseCRF with CNN into a single Recurrent Neural Network (RNN) to transform the DenseCRF post-processing
    into an end-to-end procedure.\\

 \indent Although CNN is a powerful tool for semantic segmentation,
 there are some technical hurdles when the existing CNN architectures are applied to a pixel-wise
prediction for face parsing. First, the diverse, contextual and mutual relationships among the key components for face parsing should be well addressed when predicting pixel-wise labels. Second, the predicted label maps are desired to be detail-preserved and
of high-resolution in order to recognize or highlight very small labels (e.g. the eyebow regions). However,
most of the previous works on semantic segmentation with a CNN can only predict the labels of very
low-resolution pixels, for example, the eight-time down-sampled features in the fully convolutional network (FCN) \cite{FCN,DEEPV1}.
Their prediction is very coarse and not optimal for the required fine-grained segmentation.
Third, the critical segmentation-specific context constraints, such as the local superpixel smoothness or the integrity
and uniqueness for each semantic region, have not been well considered in the previous works on face parsing.
For instance, the pixels within the same superpixel or neighboring superpixels should have
high possibilities to be assigned with the same semantic label. The label probabilities from neighboring
superpixels should help guide the label inference by leveraging the location priors. Furthermore, the pixels
within the same semantic region (e.g. eye region) should be predicted to have the same semantic label
to retain the region integrity. Lastly, although the structured prediction is a powerful tool for improving segmentation, the
training and inference computation cost of structured prediction is expensive.\\

 \indent In this paper, we present a novel fully-convolutional continuous CRF neural
 network that successfully addresses the above-mentioned issues. FC-CNN aims to capture
 cross-layer context and local super-pixel context by combining three sub-networks: a unary network, a pairwise network and a continuous CRF network. Firstly, to recover the image details, a carefully designed unary network which is composed of convolutional blocks and deconvolutional blocks is proposed. To recover the image details, we apply deconvolution layers which are trained in an end-to-end way to up-sample the feature maps layer by layer. Secondly, the within-superpixel smoothing and cross-superpixel neighborhood relationship are leveraged to retain the local boundaries and label consistencies within the super-pixels. They are formulated as natural
sub-components of the FC-CNN in both the training and the testing process. Thirdly, a pairwise network is designed to learn the pixel-wise affinity so as to capture the spatial relationship between superpixels. Finally, to incorporate the output of unary network and pairwise network into a unified framework, a particular type of graphical model, the continuous Conditional Random Field is used, which allows
us to perform exact and efficient Maximum-A-Posteriori (MAP) inference. A differentiable superpixel-pooling layer and a continuous CRF layer are designed to combine the unary network and pairwise network. Even though Conditional Random Fields are
 unimodal and as such less expressive, Continuous Conditional Random Fields are unimodal conditioned on the data,
effectively reflecting the fact that given the image one solution dominates the posterior distribution. The
C-CRF model thus allows us to construct rich expressive structured prediction models that still lend
 themselves to efficient inference. To solve the Gaussian CRF effectively, we apply very efficient algorithms
for inference and learning, as well as a customized technique adapted to the semantic segmentation task
building on standard tools from numerical analysis. Our contributions are summarized as follows:\\
      (1) We propose a deep fully-convolutional continuous CRF network which is composed of a unary network, a pairwise network and a continuous CRF network. The proposed architecture is able to integrate superpixel content, semantic edge information and continuous CRF
       model into a unified framework effectively.\\
      (2) We introduce a pairwise network to learn the pixel-wise similarity and a superpixel-pooling layer is designed to enable a fully end-to-end training of the proposed semantic segmentation network based on superpixels.\\
      (3) We present a continuous CRF layer that are designed through the solutions of linear systems.\\
      (4) We compare FC-CNN with the state-of-the-art results on the LFW-PL and the Helen datasets. Better segmentation performances in terms of class average accuracy are achieved.\\

  \begin{figure*}[htbp]
  \begin{center}
   \centerline{\includegraphics[width=15cm,height=8cm]{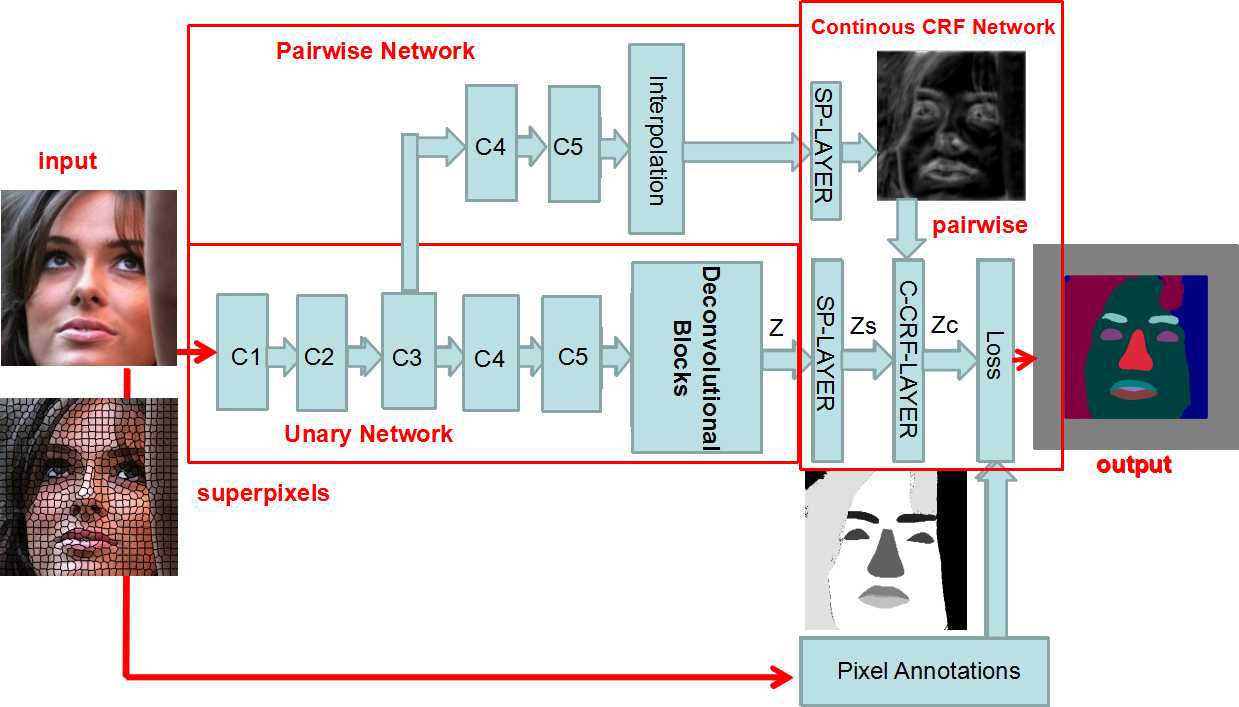}}
  \caption{The flowchart of the proposed deep network for face parsing.}
  \end{center}
 \label{fig:fw}
 \end{figure*}

\section{related work}
\subsection{Face Parsing}
 The task of face parsing is to parse an input face image into semantic regions,
 e.g, eyes, eyebrows and mouth for further processing. Face parsing provides a
 robust representation by assigning a semantic label to every pixel of a face image.
 Recently, researchers have proposed several face parsing algorithms \cite{MOFL,HFACE,CRFFACE1,CRFFACE2,EXEMPLAR}. The first category is deep learning based methods. Liu et al. \cite{MOFL} proposed a deep convolutional network that jointly models pixel-wise likelihoods and label dependencies through a multi-objective learning method. In \cite{HFACE}, Luo et al. proposed a face parsing method based on deep hierarchical features and several trained models. The second category is CRF based model. Warrell and Prince \cite{CRFFACE1} used a CRF for labeling facial components by combining a family of multinomial priors to model facial structures. In \cite{CRFFACE2}, Kae et al. modeled the face shape prior with a restricted Boltzmann machine and combined it with a CRF model for labeling threes classes (face, hair and background) of pixels. The third category is exemplar based methods. In \cite{EXEMPLAR}, Smith et al. developed a method based on transferring labeling masks from registered exemplars to classification probabilities to labeling facial skin and facial components such as nose, eyes, etc.\\

\subsection{Semantic Segmentation}
\indent In recent years, Deep Convolutional Neural Networks (CNNs) \cite{CNNS,alexnet} have demonstrated their excellent performance for semantic segmentation \cite{SDS,FCN,decov,SEGNET,DEEPV1,DEEPLABDT}. In \cite{SDS}, CNN features are applied to classify each region into one of the semantic classes. Different from the region based approaches, FCN shown in \cite{FCN} applies the full convolution only once on an entire image to directly extract features at each pixel. However, the output of FCNs tends to have poorly localized object boundaries due to the deployment of max-pooling layers and down sampling. Several approaches have been introduced to handle this problem. The methods shown in \cite{FCN} proposes to extract features from the intermediate layers of a deep network to better estimate object boundaries and recover the image details. In \cite{FCN}, a single deconvolutional layer is added in the decoding stage to generate prediction results using stacked feature maps from intermediate layers. In \cite{decov} and \cite{SEGNET}, deconvolutional layers are constructed by mirroring the convolutional layers using the stored and pooled locations in an unpooling step.
 The deconvolutional layers and the unpooling layers are employed to recover the "spatial-invariance" effect of max-pooling layers.
 Noh et al. \cite{decov} showed that coarse-to-fine structures are crucial to recover the fine-detailed information along with the propagation of deconvolutional layers. Bilinear interpolation \cite{DEEPV1,DEEPLABDT} is also commonly used because it is fast and memory-efficient.
Approaches shown in \cite{farabet2013learning,feedforward,2016superpixel} use the superpixel representation, which is essentially generated by the low-level image segmentation methods to improve the localization and segmentation accuracy.\\
\indent Although CNNs have been shown to work very well for semantic segmentation,
they may not be optimal as they can not model the interactions between variables. Combining the
strength of CNNs and CRFs for segmentation is another way to recover the fine-detailed information in an image
, and this has been the focus of recently developed approaches. Deeplab-CRF shown in \cite{DEEPV1} trains an FCN and uses a
dense CRF \cite{fullyc} in a post-processing step to refine the object boundaries by leveraging the color contrast information. CRF-RNN \cite{CRFFCN} implements recurrent layers for end-to-end learning of the dense CRF and the FCN network. It uses Potts-model based pairwise potential functions to enforce smoothness only. Lin et al. \cite{LIN} proposed a method that combined CNNs and
CRFs to exploit complex contextual information for semantic image segmentation. The CNN based pairwise
potentials are formulated for modeling the semantic relations between image regions. In \cite{gaussianCRF}, an end-to-end
trainable Gaussian Conditional Random Field network, which unfolds a fixed number of Gaussian mean field inference steps is proposed.
In \cite{DEEPRFT}, a structured prediction technique that combines the virtues of Gaussian Conditional Random Fields  with a Deep
Learning is proposed which learns features and model parameters simultaneously in an end-to-end FCN training algorithm.\\
\indent In contrast to the work described above, our approach
shows that it is more efficient to perform face parsing by designing an architecture that
integrates fully-convolutional layers, deconvolutional layers, superpixel information, continuous Conditional Random Field model and semantic edge context into a unified framework.\\

\section{The Proposed Network Architecture}
\indent Figure \ref{fig:fw} displays the flowchart of the proposed deep neural network for face parsing at a higher level.
The proposed architecture is composed of three parts: a unary network, a pairwise network and a continuous CRF network.
The unary network includes convolutional blocks and its corresponding deconvolutional blocks. The convolutional
blocks are designed to transform an input image to multidimensional feature representations.
The deconvolutional blocks are applied to recover the pixel level prediction information from the features
extracted by the convolutional layers. The pairwise network which is fed into the continuous CRF network is used to learn the similarity between pixels.
The continuous CRF network is composed of a superpixel-pooling layer, a continous CRF layer and a final softmax classification layer. The role of superpixel-pooling layer is to transform the pixel level features to superpixel level features. The softmax classification laye is to generate a probability map related to predefined classes. \\

\subsection{Unary Network}
%
 \begin{figure*}[htbp]
   \begin{center}
   \centerline{\includegraphics[width=13cm,height=5cm]{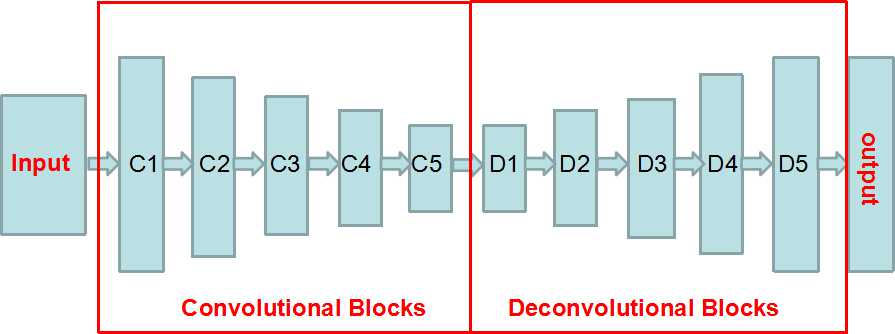}}
 \end{center}
  \caption{The architecture of the unary network.}
 \label{fig:unetv1}
 \end{figure*}
 \indent We formulate the unary function of CRF by stacking the unary network for generating the feature maps and a
 fully convolutional network to generate the final output for the unary potential function. The proposed unary network structure encodes the local details in an early stage. Different spatial resolutions are used for capturing different levels of semantic information.  The "unary" part of our network (illustrated in Figure~\ref{fig:unetv1}) is built on top of SEGNET architecture \cite{SEGNET} and we extend it to address the task of face parsing. The convolutional part computes a convolutional feature map $\tilde{Z}=f_{cov}(x)$ of input image $x$. It is initialized from the Imagenet-trained VGG-16 network \cite{verydeepcnn} and then fine-tuned on a face segmentation dataset. Each convolutional block (C1-C5) performs convolution with a filter bank to produce a set of feature maps. To reduce the internal-covariate-shift problem, a batch normalization \cite{ioffe2015batch} layer is added to the output
of every convolutional layer. Then an element-wise rectified linear non-linearity (ReLU) is applied. After that, max-pooling with a $2\times 2$ window and stride 2 is performed and the resulting output is sub-sampled by a factor of 2. \\
 \indent The feature maps from deep layers often focus on global structure and are insensitive to local boundaries and spatial displacements. It is observed that the activation maps obtained by the convolutional layers are not sufficient for semantic segmentation
since they assign high scores to only few discriminative parts of an object and their resolution are too low to recover object shape accurately.
We thus add a non-linear deconvolution module $Z=f_{dconv}(\tilde{Z})$to recover the image details from $\tilde{Z}$ generated by convolutional blocks. This module consists of five deconvolution blocks (D1-D5). In each deconvolution block, a unpooling layer is employed to reconstruct the original size of activations. The unpooling operation is applied for upsampling the feature maps \cite{decov,SEGNET}. The locations of maximum activations selected during pooling operation the input feature maps are recorded and the activations are upsampled using the memorized max-pooling indices from the corresponding convolutional activations. The output of an unpooling layer is an enlarged activation map which is then convolved with a trainable convolutional filter bank to produce dense feature maps. Then each convolutional filter bank is followed by a batch normalization layer and a rectified linear unit in deconvolution blocks\cite{SEGNET}. To recover the features effectively, a hierarchical structure of deconvolutional blocks (D1-D5) are used to recover the image details layer by layer. The filters in lower layers tend to capture overall shape of an object while the class-specific fine-details are encoded in the filters in higher layers\cite{decov}. \\


\subsection{Pairwise Network}
\begin{figure*}[htbp]
   \begin{center}
   \centerline{\includegraphics[width=14cm,height=5cm]{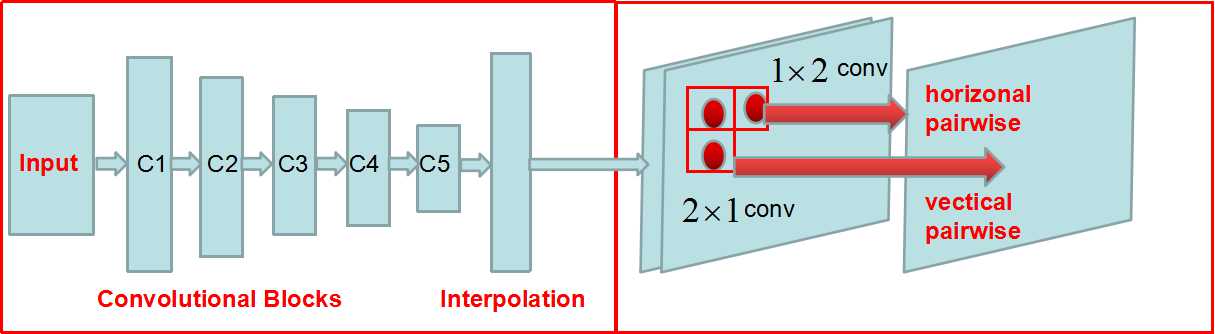}}
 \end{center}
  \caption{The architecture of the pairwise network.}
 \label{fig:pnet}
 \end{figure*}

\indent The pairwise network is designed to learn the pixelwise similarity. As illustrated in Figure \ref{fig:pnet}, a new branch of convolutional blocks C1-C5 are used to generate feature maps. Then a interpolation layer is applied to recover the feature maps to the same resolution with the original image. To compute the pairwise similarity, we create a similarity graph and each location in the feature map (which corresponds to a pixel in the input image) corresponds to a node in the graph. Pairwise connections in the pixel graph are constructed by connecting one node to its neighboring node. We consider two kinds of spatial relations by defining horizonal and vertical relations connections, and each type of spatial relation is modeled by a specific pairwise potential function. Hence the pairwise network generates a similarity matrix $WP_{ij}$ between pixels $i$ and $j$ based on the learned horizonal and vertical relations.\\
\indent The edge features are computed by concatenating the corresponding feature vectors of two connected nodes (similar to \cite{LIN}).
 The edge feature of pixel pair is then sent to a convolutional layer to compute horizonal and vertical relations. In our experiment, we use a $1*2$ convolution kernel and a $2*1$ convolution kernel to learn the horizonal and vertical pairwise relations respectively. Then the output of pairwise net is fed into the continuous CRF network. Note that the first three convolutional blocks $C1-C3$ are shared with the unary network, then two convolutional blocks are followed. Finally, a interpolation layer is presented in where the detailed information is recovered by interpolation operation.\\

\subsection{Continuous CRF Network} \label{sec:3}
\begin{figure*}[htbp]
  \begin{center}
   \centerline{\includegraphics[width=16cm,height=10cm]{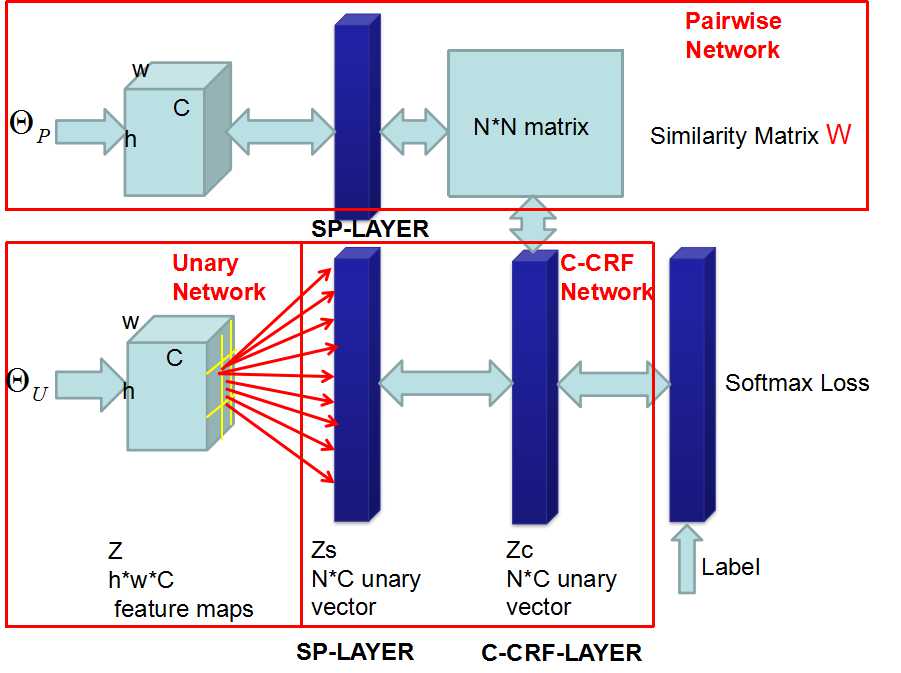}}
 \end{center}
  \caption{The details of the  continuous CRF network architecture. w represents the image width, h represents the image height and C stands for the feature channels. Z is the output of unary network. $Z_s$ is the output of SP-LAYER corresponding to unary network, $Z_c$ is the output of the CRF layer.}
 \label{fig:c3}
 \end{figure*}

\indent As shown in Figure \ref{fig:fw}, the continuous CRF network is an important part in the proposed architecture.
 The output of unary network and pairwise network are taken as the input of continuous CRF network. Two novel layers, superpixel-pooling layer (SP-LAYER) and continuous CRF layer (C-CRF layer) are designed for the continuous CRF network. The superpixel-pooling layer is designed to transform the pixel level feature representations to superpixel level feature representations. The continuous CRF model is integrated into the whole framework via the C-CRF layer. The architecture details of the proposed SP-LAYER and C-CRF layer are illustrated in Figure \ref{fig:c3}.\\
\indent \textbf{Superpixel-pooling layer (SP-LAYER):}
 Unlike the traditional practice of treating the complex superpixel random field regularization as post-processing \cite{W19}, we embed
the within-superpixel smoothing into the training stage and the testing stage. Before the C-CRF layer
 of C-CRF network, the within-superpixel smoothing is designed to project the pixel level features to region level features.
 Unlike traditional pooling layers, the pooling layout of the SP-LAYER is not pre-defined but determined by superpixels of the input image.
  Through the SP-LAYER, we can aggregate feature vectors spatially aligned with superpixels by average pooling. To simplify the notations, we assume the image is divided into $N$ superpixles after the oversegmentation step. The information from SP-LAYER will be propagated through the C-CRF layer later.\\
   \indent The roles of SP-LAYER can be summarized as two parts. The first role is to process the output of pariwise network.
  Based on the pixel level similarity $WP_{ij}$ learnt from the pairwise network, we compute the superpixel level similarity by projecting pixel level similarity to superpixel level. Then the similarity between superpixels $p$ and $q$ is defined as:
  \begin{equation}
     W_{pq}=\frac{1}{|S_p \bigcap S_q|} \sum_{i,j \in S_p \bigcap S_q} WP_{ij}.
  \end{equation}
where $S_p$ denotes the set of pixels in superpixel $p$, $S_p \bigcap S_q$ represents the boundary pixels between $S_p$ and $S_q$. \\
 \indent The second role of SP-LAYER is to process the output of the unary network. Let $Z_s$ stand for the transformed output of unary network through SP-LAYER (Figure \ref{fig:c3}). The smoothed confidence map $Z_s$ of superpixel $p$ can by represented by:
      \begin{equation}
         Z_s(p)=\frac{1}{\| S_p \|} \sum_{i \in S_p} Z(i)
      \end{equation}
 where $i$ is the index of pixels in $S_p$, $Z$ represents the output unary network. Then the output of superpixels pooling layer
 can be represented by a vector $Z_s=[Z_s(1),...,Z_s(N)]$. The output of SP-LAYER then becomes an $N \times C$ matrix, where $N$ means the numbers of superpixels and $C$ indicates the number of channels in the feature maps. \\

\indent \textbf{Continuous CRF layer (C-CRF-LAYER):}
\indent The output of SP-LAYER corresponding to unary network $Z_s$ and the output of SP-LAYER corresponding to pairwise network $W$ are fed into the C-CRF layer which implements the continuous condition random field model for segmentation performance improvement. Let $Z_c$ denote the output of $C-CRF$ layer. The unary potential of C-CRF is constructed from the output of a SP-LAYER $Z_s$ by considering the least square loss:
  \begin{equation}
      U(Z_c,Z_s;\theta_u)=\lambda(Z_c(p)-Z_s(p)(\Theta_u))^2, \forall p =1,...,N.
  \end{equation}
  where $\lambda$ is the weighting coefficient and $\Theta_u$ represents the parameters for unary network. The pairwise potentials are defined as:
   \begin{equation}
      V(Z_c(p),Z_c(q);W)=W_{pq}(Z_c(p)-Z_c(q))^2,
  \end{equation}
where $W_{pq}$ is the pairwise part from the superpixels pair $(p,q)$ which are learnt from the pairwise network.
With the unary and pairwise potentials defined, we can now rewrite the energy function:
  \begin{equation}
  \begin{aligned}
  E(Z_c,Z_s)=&\sum_{p=1}^N \lambda (Z_c(p)-Z_s(p)(\Theta_u))^2+\\
     & \sum_{(p,q) \in NE}W_{pq}(Z_c(p)-Z_c(q))^2,
      \end{aligned}
  \end{equation}
where $NE$ is the set of adjacent superpixels. To simplify the equations, the following notation is presented:\\
  \begin{equation}
      \Phi=D-W,
  \end{equation}
where $I$ is the $N\times N$ identify matrix. $W$ is the superpixel similarity matrix which is comprised of $W_{pq}$, and $D$ is
the diagonal matrix with $D_{pp}=\sum_{q}W_{pq}$. Then we have:
  \begin{equation}\label{eq:1}
      E(Z_c,Z_s)=\frac{1}{2}Z_c^T(\Phi+\lambda I)Z_c-Z_c^TZ_s+\frac{1}{2} Z_s^T Z_s.
  \end{equation}
 Given $Z_s$ and $\Phi$, the inference involves solving for the value of $Z_c$ that minimizes the energy function in Eq.\ref{eq:1}.
   \begin{equation}\label{eq:linear}
     Z_c=(\Phi+\lambda I)^{-1}Z_s.
  \end{equation}

 To solve the linear system of Eq.\ref{eq:linear}, the sequential mean-field method using the Gauss-Seidel
algorithm \cite{press1982numerical} is applied. Then the high dimensional feature representation at the output of the C-CRF layer is fed to a trainable soft-max classifier. This soft-max layer classifies each pixel independently. The output of the soft-max classifier is a $K$ channel feature map of probabilities where $K$ is the number of features. The predicted segmentation corresponds to the class with maximum probability at each pixel.
 \section{End-to-end Training for Face Parsing}
 \indent In this section, we will describe the way for training the network in an end-to-end way. The method for training the continuous CRF layer, the superpixel pooling layer, the unary network and the pairwise network will be introduced.
 \subsection{Training on Continuous CRF Network}
 \indent Firstly, we will introduce the way for computing derivatives of loss with respect to $Z_c$ and $Z_s$. Note that $Z_c$ is the output of C-CRF layer and $Z_s$ is the output of SP-LAYER related to unary network. As shown in Fig \ref{fig:c3}, the continuous CRF layer is connected to the softmax loss layer. When the proposed network is trained, the derivative $\frac{\partial L}{\partial Z_c}$ is back propagated from the above softmax loss layer.
  The SP-LAYER related to the unary network is connected to the C-CRF layer, the derivative of loss with respect to $Z_s$ is computed using the following chain rule:
  \begin{equation}\label{cr1}
   \frac{\partial L}{\partial Z_s}=\frac{\partial L}{\partial Z_c}\frac{\partial Z_c}{\partial Z_s}.
   \end{equation}
  Based on Eq (\ref{eq:linear}), the application of chain rule (Eq (\ref{cr1})) yields a closed form expression, which is a system of linear equations:
  \begin{equation}\label{eq:tr}
   \frac{\partial L}{\partial Z_s}=(\Phi+\lambda I)^{-1} \frac{\partial L}{\partial Z_c}.
  \end{equation}
  Based on the above equations, the expression for the partial derivatives of $\Phi$ is derived by using then following chain rule of differentiation:
\begin{equation}
  \frac{\partial L}{\partial \Phi}=\frac{\partial L}{\partial Z_c} \frac{\partial Z_c}{\partial \Phi}.
\end{equation}
Using the expression of calculating the matrix derivative we have:
\begin{equation}
 \begin{aligned}
    \frac{\partial Z_c}{\partial \Phi}= \frac{\partial (\Phi+\lambda I)^{-1} Z_s}{\partial \Phi}=& -(\Phi+\lambda I)^{-T}\bigotimes \\
    &(\Phi+\lambda I)^{-1}Z_s,
 \end{aligned}
\end{equation}
where $\bigotimes$ denotes the kronecker product. Then the following expression can be obtained:
\begin{equation}\label{eq12}
    \frac{\partial L}{\partial \Phi}=-\frac{\partial L}{\partial Z_s} \bigotimes Z_c.
\end{equation}

\subsection{Training on Unary Network}
  \indent To determine the partial derivatives through the superpixel-pooling layers, we observe that the superpixel pooling layer does not have any weights and we only need to compute the subgradients of the loss with respect to the pixel-level scores. We further integrate the within-superpixel smoothing into the training and testing process to utilize the detailed information. The superpixel guidance is only
  used to process the output of unary networks and pairwise network instead of all covolutional layers, so as not to influence the learning of convolution filters.\\
  \indent  In the backward process, the output of the unary network is connected to SP-LAYER. The derivative $\frac{\partial L}{\partial Z_s}$ is back propagated through the SP-LAYER to the unary network. Then, the loss derivation related to $Z$ can be computed using the following rule:
  \begin{equation}
    \frac{\partial L}{\partial Z(i)}=\frac{\partial L}{\partial Z_s(p)},
  \end{equation}
   where $i \in S_p$ that means that pixel $i$ is contained in superpixel $p$. \\
Then we have a loss tensor $ \frac{\partial L}{\partial Z}$ with dimensions  $H\times W\times C$. The network parameters of unary network
$\Theta_u$ can be trained by SGD based on back propagation chain rule in an end-to-end way:
   \begin{equation}
      \frac{\partial L}{\partial \Theta_u} =\frac{\partial L}{\partial Z} \frac{\partial Z}{\partial \Theta_u}.
  \end{equation}

\subsection{Training on Pairwise Network}
The pairwise term is learnt through the pairwise branch. The derivatives of loss with respect to similarity matrix $W$ is
defined as:
 \begin{equation}
    \frac{\partial L}{\partial W}=\frac{\partial L}{\partial \Phi} \frac{\partial \Phi}{\partial W}.
\end{equation}
 The derivates of loss $L$ with respect to $\Phi$ is defined in equation (\ref{eq12}), and the derivates $\frac{\partial \Phi}{\partial W}$ can be obtained based on the equation $\Phi=D-W$. Then the rule for back propagating gradients through the SP-LAYER related to pairwise network is defined as:
 \begin{equation}
 \frac{\partial L}{\partial WP_{ij}}= \frac{1}{|\Pi|} \sum_{i,j \in \Pi} \frac{\partial L}{\partial W_{p,q}},
  \end{equation}
 where $\Pi$ is the set of all edges which include the pixel with location $(i,j)$. In the pairwise network, the pairwise similarities are
 learnt through $1\times 2$ and $2\times1$ convolutions. Then the derivatives $ \frac{\partial L}{\partial WP_{ij}}$ are back propagated to the pairwise net to learn the pairwise network parameters $\Theta_p$.\\

\section{Experimental Results}\label{sec:5}
\indent We perform a thorough comparison of FC-CNN to the state of the art along with comprehensive
experiments. We use the LFW-PL dataset \cite{CRFFACE2} and  HELEN  dataset \cite{EXEMPLAR} to evaluate the proposed method.
We report the F-measure metric \cite{MOFL,SMITH} to measure the per-pixel segmentation accuracy. \\

\subsection{Datasets and Setting}
\indent The LFW-PL dataset \cite{CRFFACE2} contains 2927 face images of $250\times 250$
pixels acquired in unconstrained environments. All of them are manually annotated with skin, hair
and background labels using superpixels. The training set of this dataset is composed of
1500 images, the testing set is composed of 927 images, and a
validation set with 500 images is contained as well. The HELEN  dataset \cite{EXEMPLAR} contains face labels
with 11 classes for the second set of experiments. It is composed of 2330 face images of $400\times 400$ pixels with labeled
facial components generated through manually-annotated contours along eyes, eyebrows, nose, lips and jawline. The dataset is
also divided into a training set with 2000 images, a testing set with 100
images, and a validation set  with 300 images.\\

\subsection{Implementation}
\indent  The high-level architecture of our system is implemented using the popular Caffe \cite{jia2014caffe} deep learning library. We initialized the convolutional blocks and deconvolutional blocks of the unary network using SegNet network from \cite{SEGNET}.  This acts like a strong baseline for a purely feedforward network. The proposed network has symmetrical configuration of convolution and deconvolution network centered around the last pooling layer.
 In the implementation of pairwise network, the convolutional blocks C1-C3 are shared with the unary stream. Then two convolutional blocks are followed and a bilinear interpolation layer is designed to upsample the activation maps.\\
 \indent The superpixel-pooling layer and continuous CRF layer are the core of our architecture. In the implementation of C-CRF layer,
 we extend the implementation of pixel based gaussian CRF layer \cite{DEEPRFT} to the superpixel based C-CRF layer by combining the information generated by SP-LAYER.\\

\subsection{Evaluation on LFW-PL Dataset}
  \begin{table*}[htb]  
\newcommand{\tabincell}[2]{\begin{tabular}{@{}#1@{}}#2\end{tabular}}
\begin{center}
\begin{tabular}{|l|l|l|l|l|}
\hline
           &{\bf F-skin} &{\bf F-hair} &{\bf F-bg} &\tabincell{l}{\bf overall \\ \bf accuracy} \\
\hline
MO-GC \cite{MOFL}  &93.93 & 80.70 &\textbf{97.10} &95.12 \\
SEGNET \cite{SEGNET} & 93.15 &84.18 &95.25 & 93.56 \\
FCN \cite{FCN}  &92.91 &82.69 &96.32 &94.13 \\
CRFASRNN \cite{CRFFCN}   &92.79 &82.75 &96.32 &94.12 \\
DEEPLAB  \cite{DEEPV1}  &92.54  &80.14 &95.65 &93.44 \\
DEEPLAB-DT \cite{DEEPLABDT} &91.17  &78.85 &94.95 &92.49 \\
FC-CNN wo superpixels    & 94.02 &84.73 &96.06 &95.06 \\
FC-CNN     & \textbf{94.10} &\textbf{85.16} &96.46 & \textbf{95.28} \\

\hline
\end{tabular}
\end{center}
\caption{Overall accuracy on LFW-PL dataset. The F-measures of skin (F-skin),
hair (F-hair) and background (F-bg) are presented.}
 \label{t:tvoi}
\end{table*}

\indent At the first step we directly compare the potential advantage of FC-CNN with respect to the state-of-that-art methods on the task of labeling facial components such as skin, hair and background. In this task, the superpixels are generated using LSC \cite{LSC}. Note that our method can be used with any oversegmentation algorithm. As a first step, we directly compare the proposed FC-CNN with current face parsing method \cite{SMITH,MOFL} and state-of-the-art semantic segmentation methods, including  FCN \cite{FCN}, CRFASRNN \cite{CRFFCN}, SEGNET \cite{SEGNET}, DEEPLAB \cite{DEEPV1}, DEEPLAB+CRF \cite{DEEPV1}, DEEPLAB-RTF \cite{DEEPRFT} and DEEPLAB-DT \cite{DEEPLABDT}.\\
  \indent In the experiments, we train our method on the training set and validation set of LFW-PL and evaluate them on the test images. All images are cropped to an input resolution of $320 \times 320$ to adapt to our network architecture. In the construction of pairwise net, each pixel is connected to its left, right, top and bottom adjacent neighbours. When FC-CNN is compared with other methods, we cropped the images so as to adapt to different kinds of deep learning models. For example, the images are padded to the size of $500 \times 500$ for FCN and CRFASRNN. For SEGNET, images are transformed to the size of $512 \times 512$ by padding the border regions with zeros. For DEEPLAB and DEEPLAB-DT, the images are cropped to the size of $353 \times 353$.\\
\indent The quantitative results of the proposed method and the competitors are presented in Table \ref{t:tvoi}. We can see that FC-CNN achieves the higher accuracy on facial skin and hair segmentation accuracy compared to the pixel classification based face parsing method MO-CG. A $2\%$ overall segmentation accuracy improvement can be achieved by FC-CNN compared with the baseline SEGNET. Compared to other fully convolutional methods, FC-CNN achieves the highest accuracy over all the three classes. \\
\indent We also evaluate the role of superpixel information and superpixel pooling layer on LFW-PL dataset. As shown in Table \ref{t:tvoi}, the integration of superpixels information can improve the segmentation accuracy. Figures will demonstrates effectiveness of pixel-wise prediction for facial parsing, and we can observe that FC-CNN shows better performance than the compared methods by recovering the detailed image information.\\

  \begin{table*}[htb]  
  \newcommand{\tabincell}[2]{\begin{tabular}{@{}#1@{}}#2\end{tabular}}
\begin{center}
\begin{tabular}{|l|l|l|l|l|l|l|l|l|l|}
\hline
Methods     &{\bf brows ~~} & {\bf eyes ~~} &{\bf nose ~~} &\tabincell{l}{\bf upper  lip} &\tabincell{l}{\bf in  mouth} &\tabincell{l}{\bf lower lip} &\tabincell{l}{\bf mouth all} &\tabincell{l}{\bf facial  skin} &{\bf overall}  \\
\hline
   MO-GC \cite{MOFL}           & 0.734  & 0.768  & 0.912  & 0.601  & \textbf{0.824}  &0.684 & 0.857 & 0.912 & 0.854 \\
   FCN \cite{FCN} & 0.677  &  0.7429   & 0.886  &  0.624   & 0.764   & 0.751 &0.719 &0.880 &0.862 \\
  CRFASRNN \cite{CRFFCN} & 0.682   &  0.769    & 0.885   &  0.627   &  0.769   &0.774 & 0.732 & 0.896  & 0.877  \\
  DEEPLAB \cite{DEEPV1} & 0.661   &  0.704  &   0.878 &    0.585 &    0.701  &0.724  & 0.678 &0.881 &0.858 \\
  DEEPLAB-DT \cite{DEEPLABDT} & 0.700  &  0.754  &  0.901  &  0.638  &  0.738 &0.762  & 0.721 &0.901 &0.880 \\
  Simth $et.al$ \cite{SMITH}    & 0.722     & 0.785  & \textbf{0.922}   & 0.651 & 0.713  &0.700 & \textbf{0.857} & 0.882 & 0.804 \\
  DEEPLAB-CRF \cite{DEEPV1} & 0.401  &   0.728  &   0.807  &   0.460 &    0.702 & 0.717  &0.643 & 0.910  & 0.871  \\
  DEEPLAB-RFT \cite{DEEPRFT} & 0.701  &  0.736  &  0.886 &   0.624  &  0.719 & 0.749 &0.706  & 0.889 &0.868 \\
  SEGNET \cite{SEGNET}   & 0.747   & 0.810   & 0.898  &   0.708   &  0.756  &0.796& 0.762 & 0.902  & 0.887  \\
  FC-CNN wo superpixel   & 0.751  & 0.819  &0.901  &  0.711  &0.793 & 0.811 &0.771  & 0.905 & 0.893  \\
  FC-CNN   & \textbf{0.757}  & \textbf{0.828}  &0.906   & \textbf{0.717} &0.799  & \textbf{0.817} &0.782  & \textbf{0.911} & \textbf{0.897} \\
 \hline

\end{tabular}
\end{center}
\caption{ Overall accuracy on HELEN dataset. The F-measures of seven categories, the mouth-all accuracy and the overall accuracy are presented.}
 \label{t:tvoi1}
\end{table*}

\subsection{Evaluation on HELEN Dataset}
\indent In the second experiment, we conducted an experiment on the HELEN dataset,
 which differs from the LFW-PL dataset. The labels of images in HELEN dataset are composed of two
eyes, two eyebrows, nose, upper and lower lips, inner mouth, facial skin and hair. Unlike LFW-PL, some facial
components are rare classes (e.g. eyes, lips, etc) in HELEN dataset. In the preprocess of our experiment, the faces are extracted
from the original images via face detection and facial landmarks detection \cite{cao2014face}. All the cropped facial images are
resized to the size of $350 \times 350$ and then padded zeros to the size of $512 \times 512$. However, all the segmentation labels are transformed to the original sizes in the evaluation process.\\
\indent In this experiment, we merge the ground truth hair label with the background to train a
7-classes network. In this way, a fair comparison with the work of \cite{MOFL} and \cite{SMITH} can be obtained.
Based on the same subset of images with the same criteria, the experiment results on the HELEN dataset are demonstrated in Table \ref{t:tvoi1}.
A large variation on F-measure with respect to each facial components can be seen. The non-deep learning based method
such as the one shown in \cite{SMITH} based on exemplar transfer obtains the better result on relative rare facial classes, such as noses. Compared with fully convolutional methods, the region classification based method \cite{MOFL} performs better on class such as "in mouth", but the F-measures for most of the classes are lower than fully convolutional methods.\\
\indent We compare the proposed FC-CNN with a number of recent fully convolutional semantic segmentation methods with competitive performance. Better segmentation performance is achieved compared with methods such as FCN, CRFFCN, DEEPLAB and DEEPLABDT.
This primarily because that the facial component regions are small, the architectures such as DEEPLAB and FCN are not specially designed for segmenting small objects. Morevoer, FC-CNN generates better segmentation performance over all the classes compared with the baseline SEGNET. We can see that the primary advantage of our model comes from delineating the objects and improving fine segmentation boundaries by the carefully designed deconvolutional layers, the integration of superpixels information and the continous CRF model. \\
 \indent We also evaluate the role of superpixels and superpixel pooling layer in improving the segmentation accuracy. The corresponding results are listed in Table \ref{t:tvoi1}. The results indicate the importance of
 superpixels for segmentation performance improvement. The segmentation accuracies for all the seven classes have been improved by
 incorporating the superpixel representation into the segmentation architecture. Consistency over regions helps to get rid of spurious regions of wrong labels. The superpixels can work as guidelines for recovering detailed image information. What's more, the computation cost of continuous CRF has decreased by inferring on the superpixels based affinity matrix.\\
\indent  The qualitative results of FC-CNN and the compared methods are presented. Overall, FC-CNN produces
fine segmentations compared to other methods, and handles small objects (such as eyes, mouth and eyebows) by integrating deconvolutional
layers and superpixel information. The methods such as FCN, DEEPLAB tend to fail in labeling too small objects due to its fixed-size receptive field. Our network generally returns the object masks that are more close to the true object boundaries. \\

%
%
%
\section{Conclusions}
\label{sec:6}
 \indent We proposed a fully-convolutional continuous CRF network for face parsing. This architecture combines the advantage of adaptive representation of superpixels context, continuous Conditional Random Field with end-to-end training directly optimized for
    semantic segmentation. We achieve this by introducing a segmentation framework that includes three sub-networks: a unary network, a pairwise network and a continuous CRF network. We apply deconvolution blocks to recover the image details in unary network. Pairwise network is designed to learn the pairwise relationship between pixels. In the continuous CRF network, a differentiable superpixel-pooling layer and a continuous CRF layer are designed to combine the unary network and pairwise network.\\
    \indent Importantly, we show that continuous CRF inference is an efficient way for utilizing the superpixel information and improving segmentation accuracy. Extensive experimental results on face parsing tasks clearly demonstrate the superiority of the proposed method over the other state-of-the-art methods on LFW-PL and HELEN datasets. In the future work, we will further extend our FC-CNN architecture for more generic image parsing tasks, e.g, scene semantic segmentation.\\

%

\ifCLASSOPTIONcaptionsoff
  \newpage
\fi

\bibliographystyle{IEEEbib}
\bibliography{saliencysegmentation}
 \end{document}